\documentclass{article}

\usepackage{arxiv}
\usepackage{amsmath}
\usepackage{caption}

\usepackage[utf8]{inputenc} 
\usepackage[T1]{fontenc}    
\usepackage{hyperref}       
\usepackage{url}            
\usepackage{booktabs}       
\usepackage{amsfonts}       
\usepackage{nicefrac}       
\usepackage{microtype}      
\usepackage{xcolor}         
\usepackage{graphicx}
\usepackage{subcaption}
\usepackage{caption}

\title{Optical Flow Regularization of Implicit Neural Representations for Video Frame Interpolation}

\author{Weihao Zhuang \thanks{Equal contribution. The order of appearance was decided by the toss of a coin.} \qquad Tristan Hascoet \footnotemark[1] 
\qquad Ryoichi Takashima \qquad Tetsuya Takiguchi \\ \\Kobe University, Japan}

\begin{document}

\maketitle

\begin{abstract}
Recent works have shown the ability of Implicit Neural Representations (INR) to carry meaningful representations of signal derivatives.
In this work, we leverage this property to perform Video Frame Interpolation (VFI)
by explicitly constraining the derivatives of the INR to satisfy the optical flow constraint equation.
We achieve state of the art VFI on limited motion ranges
using only a target video and its optical flow, without learning the interpolation operator from additional training data.
We further show that constraining the INR derivatives not only
allows to better interpolate intermediate frames
but also improves the ability of narrow networks to fit the observed frames,
which suggests potential applications to video compression and INR optimization.
\end{abstract}

\section{Introduction}

Many core concepts across the fields of signal processing are defined in terms of continuous functions and their derivatives:
surfaces are continuous manifolds in space,
motion is a rate of change in space through time, etc.
In contrast, modern digital hardware is inherently discrete:
digital sensors capture discrete observations of the world regularly sampled in time and space;
computers store and process discrete representations of signals.
In order to model continuous notions on discrete signal representations,
classical methods have used different simplifying assumptions,
often taking the form of constant first or second derivatives
of the signal between consecutive observations.
The lack of generality of any such handcrafted heuristics,
combined with the ever improving quantitative results of Machine Learning (ML) methods,
have led to the near ubiquitous use of ML in recent signal processing research.
These methods leverage large collections of data to infer statistical properties of signals instead of hand-crafted heuristics.

In computer vision, Video Frame Interpolation (VFI) is one task representative of such development.
VFI models aim to interpolate intermediate frames between the consecutive frames of a video.
To do so, most successful methods rely on the optical flow
to guide the interpolation of pixel intensities from the pixel grid of observed frames onto the pixel grid of intermediate frames.
Classical methods formulate assumptions such as constant movement or acceleration fields between consecutive frames \cite{baker2011database} \cite{barron1994performance} \cite{herbst2009occlusion}.
The value of each pixel in the inferred intermediate frame is computed by first shifting the pixel intensities of observed frames along the
optical flow directions before interpolating the shifted intensities onto the intermediate frame's pixel grid.
Such approaches suffer from the following two main limitations:

\begin{itemize}
\item The optical flow is prone to errors due to occlusions, external illumination variations, etc.
\item Assumptions of constant motion field or its derivatives do not often hold true in practice.
\end{itemize}

These limitations share a common root cause: discretization.
Indeed, both the constant brightness assumption, from which is derived the optical flow,
and assumptions of constant motion field used by the interpolation process,
only truly hold at the infinitesimal scale, for time deltas typically much smaller than those of practically used Frames Per Second (FPS).

ML approaches \cite{jiang2018super} \cite{li2020video} \cite{park2021asymmetric} \cite{park2020bmbc} \cite{lee2020adacof} have instead proposed to learn the frame interpolation operator from large video collections,
without formulating explicit assumption on the signal.
While these approaches have achieved great success in terms of benchmark performance,
they are prone to generalization errors caused by domain shifts. 
Indeed differences between the training set distribution (i.e. VFI benchmark videos)
and the target video distribution may hinder the performance of ML models, e.g.;
differences stemming from the range of motion, exposure time, FPS and blur \cite{zhang2020video}.

In the mean time, research on implicit representations seeks better discrete representations of continuous signals.
In recent years Implicit Neural Representations (INR), i.e. representing signals as Neural Networks (NN)
have been shown to offer several competitive advantages over explicit representations,
with notable early successes for 3D shape representations \cite{mildenhall2020nerf}.
Of particular interest to us is the work of SIREN \cite{sitzmann2020implicit},
in which the authors have shown that representing images using Multi Layer Perceptrons (MLP) with sine activation functions
allowed for meaningful representations of the signal derivatives.
Inspired by this work, we question whether SIREN may be used to guide the interpolation
process of VFI by controlling the exact derivatives of the signal instead of the finite differences
between consecutive discrete frames, thus avoiding the pitfalls of traditional methods due to discretization.
We do so by constraining the derivatives of SIREN representations to satisfy the optical flow constraint equation,
i.e., to be orthogonal to the video's optical flow
(which we compute using existing state-of-the-art OF models).
We find that this method outperforms most existing
machine learning-based approaches on small motion range benchmarks,
without relying on machine learning for the interpolation operator.
In this sense, our method is most similar to classical VIF approaches,
except that instead of wrapping the OF on discrete explicit frame representations,
we apply the optical flow constraint on the exact gradient of the INR.
Our method is thus not subject to any mismatch between training and test data.
Furthermore, our approach can sample any number of frame in-between the observed frames
due to the continuous nature of the representation.
In addition to its application to VFI, we also show that constraining the gradient
of the model also improves the ability of narrow MLPs to fit the signal,
suggesting potential applications in INR optimization and video compression.

\begin{figure}[h]
\centering
\includegraphics[width=0.8\textwidth]{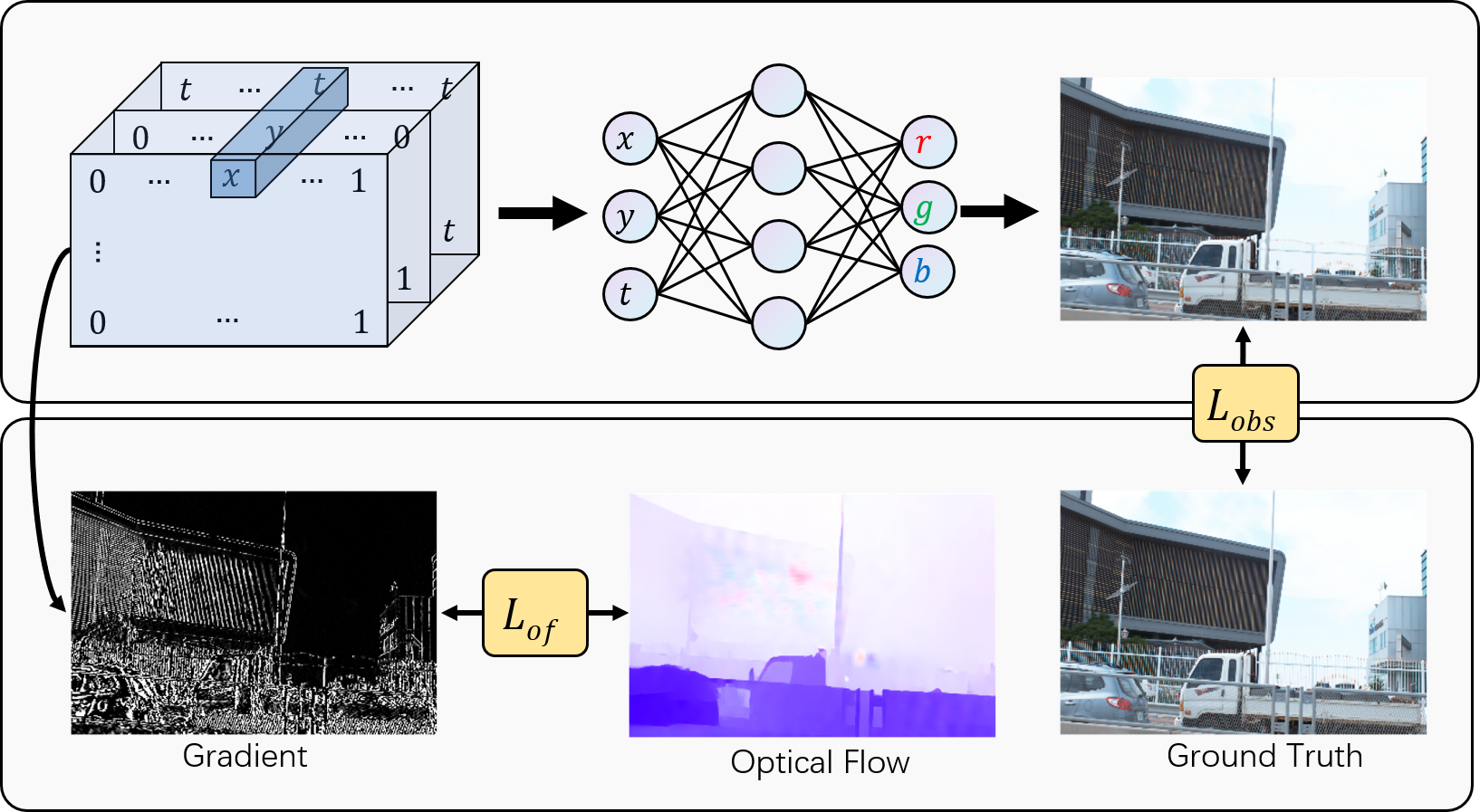}
\caption{Illustration of our approach.
We optimize SIREN to minimize the weighted sum of two losses:
The observation loss measures the fit to the video frames,
and the OF loss measures the orthogonality
between the SIREN derivatives and the video's optical flow.}
\end{figure}

To summarize the contributions of this work, we show that:
\begin{itemize}
\item SIREN representations of videos can be constrained so as to satisfy the OF constraint in their exact derivatives.
\item Such representations reach state of the art VFI on limited motion ranges, without learning a residual flow nor interpolation operator.
\item The OF constraint not only allows SIREN to generate intermediate frames, but also improve the ability of narrow SIREN to fit observed frames.
\end{itemize}

On the other hand, our approach (in its current form) presents important limitations:

\begin{itemize}
\item Optimization of the INR is very time-consuming, which hinders our ability to work on full resolution videos for time constraints.
\item Our method currently only works on limited motion range, it does not match state of the art ML models on large motion ranges.
\item It relies on an input optical flow, which is computed using existing ML-based model and is thus prone to domain shift generalization errors.
\end{itemize}


Given these limitations, the aim of this paper is not to provide a standalone production ready VFI system.
Instead, we aim to present actionable insights on a simple method
that can be either built upon or integrated to existing models.
The remainder of this paper is organized as follows:
We briefly present some related work in Section \ref{sec_related},
the detail of our method in Section \ref{sec_method},
and design several experiments to highlight the merits of our approach in Section \ref{sec_exp}.
Finally, we discuss current limitations and present potential ways to address them in Section \ref{sec_lim},
before concluding in Section \ref{sec_conc}.

\section{Related Work}
\label{sec_related}

\textbf{ Implicit Neural Representations} have met early success in
shape representation and 3D rendering \cite{park2019deepsdf} \cite{mescheder2019occupancy} \cite{mildenhall2020nerf}.
Since then, a number of works have attempted to
apply INR to different signals including audio \cite{sitzmann2020implicit} \cite{kim2022learning}, images \cite{dupont2021coin} \cite{dupont2022coin++}, videos \cite{chen2021nerv} \cite{shangguan2022learning} \cite{rho2022neural},
medical imaging and climate data \cite{dupont2022coin++}.
In \cite{sitzmann2020implicit} the authors have shown that MLP with sine activations
could fit representations of images with meaningful representations of their gradient,
and that such models could be optimized to satisfy constraints on their gradients.
Combined, these two findings have motivated our idea to apply the optical
flow constraint to the gradient of SIREN representations of videos.
A series of recent works have applied INR to video compression \cite{zhang2021implicit} \cite{chen2021nerv} ,
with some works \cite{chen2021nerv} even reporting higher PSNR
than practical codecs on high compression rates.
Although closely related to video compression,
we differ from these works as we focus on VFI.
Most related to ours is the concurrent work by Shangguan et al. \cite{shangguan2022learning}, which also uses INR for VFI.
Their approach, CURE, differs from ours in scope:
they propose to learn a prior on the INR,
while we only focus on leveraging INR to guide the interpolation process using a given optical flow.

\textbf{Video Frame Interpolation} research has largely relied on optical flow to guide the video frame interpolation process \cite{baker2011database} \cite{barron1994performance} \cite{herbst2009occlusion}.
Most works have assumed uniform optical flow between consecutive frames so as to linearly interpolate intermediate frames along the optical flow directions.
One exception is the work of \cite{xu2019quadratic}, in which the authors propose to take into account acceleration to perform the interpolation,
leading to quadratic interpolation.
Our work only constrains the first derivatives of the signal.
We differ from classical works in that we apply the OF to the exact representation derivatives,
so that we do not need to assume constancy of signal derivatives on any time interval.
Recent OF-based VFI leverages deep learning for optical flow estimation and interpolation.
Super-SloMo \cite{jiang2018super} is an important study of such methods.
The authors use a deep learning model to predict the forward and backward flows
of intermediate frames, and warp the two surrounding frames to obtain the intermediate frames.
RRIN \cite{li2020video} uses residual learning to optimize the performance of \cite{jiang2018super} at the motion estimation bound.
AMBE \cite{park2021asymmetric}, a current state-of-the-art VFI method,
proposes an asymmetric motion estimation method based on \cite{park2020bmbc},
which enhances the quality of interpolated frames by loosening the linear motion constraint.
Kernel-based approaches such as AdaCof \cite{lee2020adacof} avoid explicit separation of motion estimation and wrapping stages and
instead directly interpolate intermediate frames from consecutive observed ones.

\section{Method}
\label{sec_method}
We consider a ground-truth video as a continuous signal $v$ mapping continuous spatial ($x$, $y$) and temporal ($t$) coordinates to RGB values:

\begin{equation}
\begin{aligned}
v:& \: (x, y, t) \rightarrow (R, G, B) \\
v:& \: \mathbb{R}^3 \rightarrow \mathbb{R}^3
\end{aligned}
\end{equation}

Our goal is to find a continuous function $f_{\theta}$, parameterized by $\theta \in \Theta$,
with minimum distance $d$ to the ground-truth signal:

\begin{equation}
\begin{aligned}
f_{\theta}:& \:(x, y, t) \rightarrow (R, G, B) \\
s.t. \: \theta =& min_{\Theta} \iiint d(f_{\theta}(x,y,t), v(x,y,t)) dx dy dt
\end{aligned}
\end{equation}

where the distance function $d$ may either be the Peak Signal to Noise Ratio (PSNR) or the Structural Similarity Index Measure (SSIM).
To do so, we only have access to regularly sampled observation of the signal $v$
(i.e. the explicit representation of the video), which we denote as:

\begin{equation}
\begin{aligned}
&\mathcal{V} \in  \: \mathbb{R}^{T \times H \times W \times 3} \\
s.t. \: &\mathcal{V}_{xyt} =   v(x, y, t) 
\end{aligned}
\end{equation}

where $T$ represents the number of frames in the video, and $H \times W$ the spatial resolution.
We use SIREN as parameterized function $f_{\theta}$.
The most straightforward way to approximate Equation 2 is to optimize the model parameters so as to fit the video frames,
using the following loss function (we refer to as the observation loss):

\begin{equation}
\mathcal{L}_{obs} = \frac{1}{HWT} \sum_{x=1}^W\sum_{y=1}^H\sum_{t=1}^T || f_{\theta}(x,y,t) - \mathcal{V}_{xyt} ||^2
\end{equation}

\begin{figure}[t]
\centering
\includegraphics[width=0.8\textwidth]{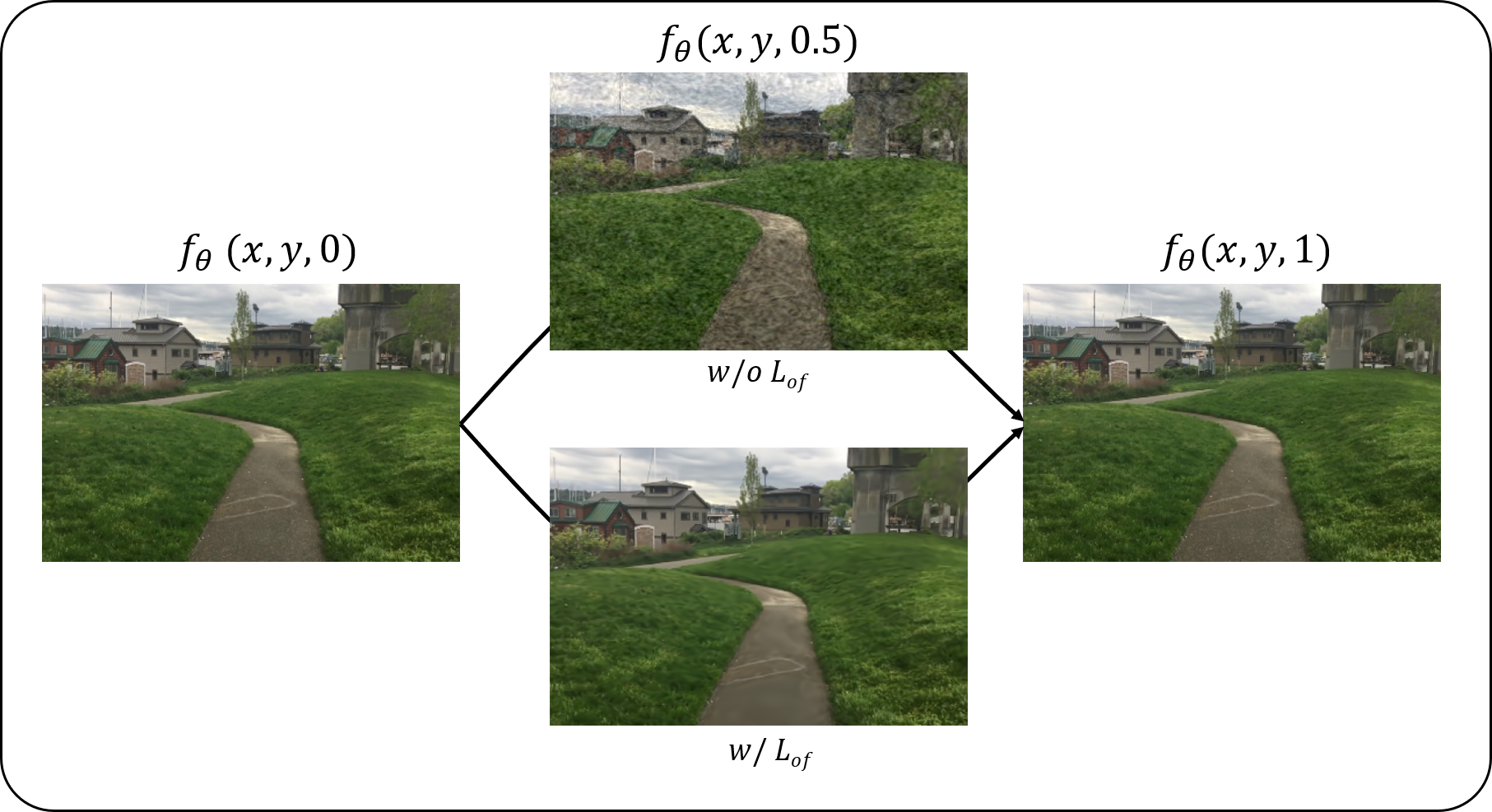}
\caption{Illustration of INR frame interpolation with and without optical flow regularization.
Without regularization (middle top), intermediate frames show unnatural high-frequency variations.
Regularizing the INR to satisfy the optical flow constraint equation result in nicely interpolated frames (middle bottom).
}
\label{fig_w_wo_OF}
\end{figure}

However, we found that optimizing the INR to only minimize this observation loss leads to overfitting the observation with high temporal frequencies:
the intra-frame signal, which we aim to correctly recover, shows important deviations from the observed frames, as illustrated in Figure \ref{fig_w_wo_OF}.
This observation has lead us to consider fitting not only the signal itself, but to also constrain its derivatives.
In particular, we regularize the model so as to respect the optical flow constraint.
The optical flow constraint equation states that for an infinitesimal lapse of time $\delta t$,
the brightness of physical points perceived by a camera at arbitrary coordinates $(x,y,t)$ should remain constant.
In other words, given the displacement $(\delta x, \delta y)$ of a physical point in the image coordinate system,
the image brightness $v$ should remain constant:
\begin{equation}
v(x, y, t)=v(x + \delta x, y + \delta y, t + \delta t)
\end{equation}

We introduce the vector notation $\textbf{x}=(x,y,t)$ for readability.
Expressing movement as a ratio of displacement in time,
we can write the optical flow $F$ and the above constraint as:

\begin{equation}
\begin{aligned}
F(\textbf{x})=(\frac{\delta x}{\delta t}, \frac{\delta y}{\delta t}, 1) \\
v(\textbf{x})=v(\textbf{x} + F(\textbf{x}))
\end{aligned}
\label{equ_6}
\end{equation}

First order Taylor expansion of Equation \ref{equ_6} gives the following
\begin{equation}
\begin{aligned}
v(\textbf{x}) = v(\textbf{x}) + \frac{\delta v}{\delta \textbf{x}} \cdot F(\textbf{x}) \\
\frac{\delta v}{\delta \textbf{x}} \cdot F(\textbf{x}) =0
\end{aligned}
\label{equ_7}
\end{equation}

which holds exactly in the limit of infinitesimal $\delta t$.
We constrain the SIREN derivatives to obey the constraint of Equation \ref{equ_7}.
Denoting the derivatives of the SIREN as:

\begin{equation}
D(f, \theta, x, y, t)=\Big(\frac{\delta f_{\theta}(x,y,t)}{\delta x}, \frac{\delta f_{\theta}(x,y,t)}{\delta y}, \frac{\delta f_{\theta}(x,y,t)}{\delta t}\Big)
\end{equation}

we can now define the optical flow regularization loss

\begin{equation}
\mathcal{L}_{of} = \frac{1}{HWT} \sum_{x=1}^W\sum_{y=1}^H\sum_{t=1}^T | D(f, \theta, x, y, t) \cdot F(x, y, t) |
\end{equation}

This loss constrains the derivatives of the signal to be orthogonal to the optical flow and
can be understood as keeping constant brightness along the optical flow directions.
The total loss we use to optimize the INR is a weighted sum of these two terms:

\begin{equation}
\mathcal{L} = (1-\lambda) \mathcal{L}_{obs} + \lambda \mathcal{L}_{of}
\end{equation}

where $\lambda$ is a hyper-parameter taking values between 0 and 1, whose impact we investigate in the following section.
The exactness of the optical flow constraint at the infinitesimal scale plays in our favor:
As we regularize the true derivative of the signal representation,
we do not assume constant motion on any time interval.
We believe this to be the main factor behind our positive results.
On the other hand, the optical flow we use was estimated from discrete consecutive frames,
and thus does not represent the true infinitesimal motion field but an estimation of finite differences.
We discuss potential alternatives in Section \ref{sec_lim}.

\section{Experiments}
\label{sec_exp}

Following previous works, we use the Adobe\cite{su2017deep}, X4K\cite{sim2021xvfi} and ND Scene\cite{yoon2020novel}
datasets as benchmark to compare our method to state-of-the-art models.
We use every two frames of each video as observations,
and evaluate the ability of SIREN to interpolate on every other (intermediate) frame.
For the Adobe dataset, we evaluate our method on the eight videos test split proposed in previous works \cite{jiang2018super}.
We run all additional experiments on the 720p240fps1.mov video of the Adobe dataset (illustrated in Figure \ref{fig_w_wo_OF}).
Due to the time-consuming operation of optimizing SIREN representations,
we optimize and evaluate all models on a $240 \times 360$ pixel resolution,
and we restrict the Adobe dataset videos to their first 40 frames.
Unless specified otherwise, we use the following default parameters:
SIREN model with depth 9, width 512 and an $\omega$ of 30.
We optimize the models with the Adam optimizer using a cosine learning rate with maximum learning rate of $10^{-5}$ during 5000 epochs.
We uses $\lambda = 0.12$ for the loss function. We compute the optical flow of videos in original resolution using the GMA \cite{jiang2021learning} OF model.

In Section \ref{sec_of_high}, we start by highlighting a trade-off akin to
underfitting vs overfitting of the signal high frequencies in vanilla SIREN representations.
We show that OF-regularized SIREN outperform the best performing vanilla SIREN,
showing that the impact of our proposed OF regularization goes beyond high frequency regularization.
In Section \ref{sec_sota}, we quantitatively compare our method to state of the art models on standard datasets.
We show that our method achieves state-of-the-art results on videos with limited motion ranges,
but underperforms recent methods for videos with large motion ranges.
We present an ablation study in Section \ref{sec_abl},
providing insights and appropriate settings for the main hyper-parameters,
and a qualitative analysis of our results in Section \ref{sec_qua}.
Finally, Section \ref{sec_video_fit} presents a surprising and counter-intuitive result:
we show that our OF loss helps SIREN converge to higher PSNR on the observed frames,
opening new potential perspectives for INR optimization and video compressions.

\subsection{Optical flow constraint and signal frequencies}
\label{sec_of_high}
\begin{figure}[h]
\centering
\includegraphics[width=0.5\textwidth]{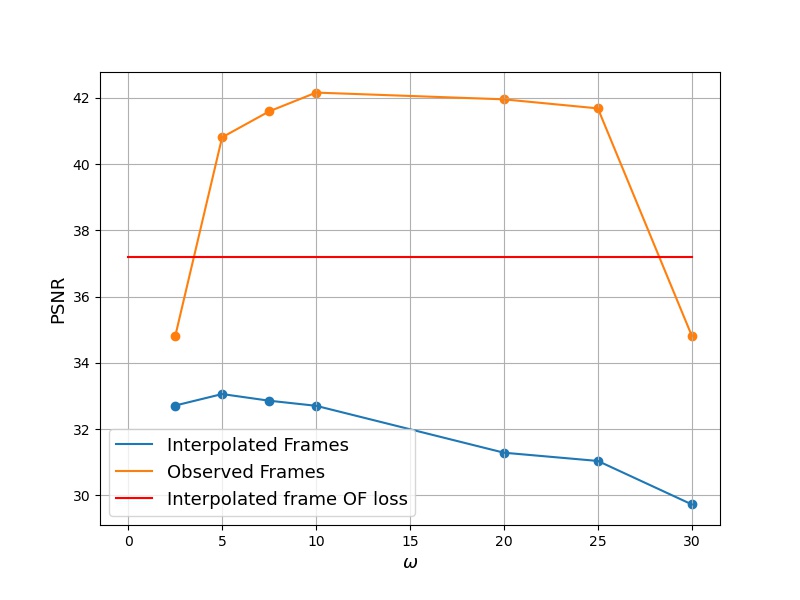}
\caption{Evolution of the PSNR of observed and interpolated frames with $\omega$ without OF loss.
Limiting the high frequency fit alone does not reach the same interpolation accuracy as the OF loss.}
\label{fig_omega}
\end{figure}

Figure \ref{fig_w_wo_OF} illustrates the fact that the OF constraint smooths
out high-frequency noise in the interpolated frames of vanilla SIREN representations.
Healthy skepticism leads us to question whether the impact of the OF constraint is limited to dampening
high frequency components of vanilla SIREN representations.
To do so, we analyze the representations of vanilla SIREN geared towards different frequency ranges,
and compare them to OF-constrained SIREN representations.
We constrain the vanilla SIREN frequencies by varying their $\omega$ parameter,
and report our comparison in Figure \ref{fig_omega},
with low $\omega$ values corresponding to lower frequency ranges.

Constraining the frequency range of vanilla SIREN with $\omega$ down to 5 degrades the fit to observed frames but improves interpolation.
This suggests that $\omega$ behaves similarly to a regularization parameter by
controlling a regime of overfitting to the observed frames high frequencies (for high $\omega$ values),
versus underfitting (for low $\omega$ values).
Figure \ref{fig_omega} further shows that OF-constrained SIREN achieve far higher interpolation PSNR than
the best performing vanilla SIREN, confirming that the OF constraint impact goes beyond dampening of the high frequency noise.
Note that we did not vary the $\omega$ of the OF constrained SIREN in this figure in order to better illustrate our point,
the red line represents results for the best performing $\omega$.
The impact of the $\omega$ parameter on OF-constrained SIREN is illustrated separately in Figure \ref{fig_abl_omega}.

\subsection{State of the art models}
\label{sec_sota}

\begin{table}[!t]
    \caption{Quantitative comparison to state-of-the-art VFI on Standard benchmarks. Results are formatted as PSNR / SSMI.}
    \begin{minipage}[t]{.5\linewidth}
      \centering
      \caption*{(a) Limited motion range}
      \begin{tabular}[t]{l | l | l }
       &  Adobe-240FPS &  X4K  \\
      \hline
      Super-SloMo \cite{jiang2018super} &  27.77 / 0.886 & 27.38 / 0.852  \\
      RRIN \cite{li2020video}  & 32.37 / 0.962 & 30.70 / 0.927  \\
      BMBC \cite{park2020bmbc}  & 27.83 / 0.917 & 27.42 / 0.858   \\
      AdaCof \cite{lee2020adacof} & 35.50 / 0.968 & 34.61 / 0.921 \\
      ABME   \cite{park2021asymmetric} & 35.28 / 0.966 & 34.30 / 0.919 \\
      FILM   \cite{reda2022film} &	35.97 / 0.971 & \textbf{35.14} / 0.939 \\
      Ours	& \textbf{36.52} / \textbf{0.977} & 35.06/ \textbf{0.944} \\
      \end{tabular}
    \label{tab_a}
    \end{minipage}%
    \begin{minipage}[t]{.5\linewidth}
      \centering
      \caption*{(b) Large motion range}
        \begin{tabular}[t]{l | l }
        	    &   ND Scene  \\
        \hline
        V-NF \cite{mildenhall2020nerf}   &  23.30 / 0.726 \\
        NSFF \cite{li2021neural}   & 28.03 / 0.925 \\
        CURE \cite{shangguan2022learning}   & \textbf{36.91} / \textbf{0.984} \\
        Ours	     & 29.22 / 0.921
        \end{tabular}
	\label{tab_b}
    \end{minipage}

\end{table}

Table \ref{tab_a} quantitatively compares the results of our method to state-of the art VFI models on different datasets.
Despite its simplicity, and without any training data, our method outperforms most existing models on limited motion ranges (Table \ref{tab_b}).
However, as illustrated in Figure \ref{fig_large}, it falls short of state-of-the-art methods on the more complex ND Scene benchmark due to larger motion ranges.
We provide further comparison in the qualitative analysis of Section \ref{sec_qua} and Section \ref{sec_lim}
discusses possible ways forward to bridging the gap performance on large motion datasets.

\subsection{Ablation study}
\label{sec_abl}

\begin{figure}[h]
\centering
\begin{subfigure}{0.3\textwidth}
	\centering
    \includegraphics[width=1\linewidth]{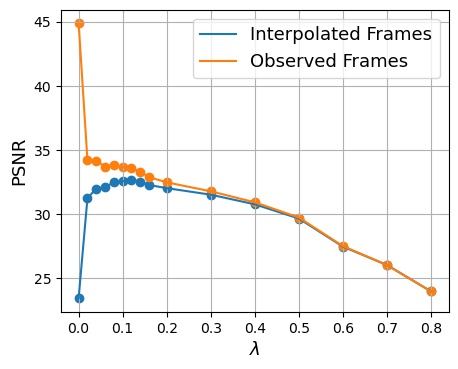}
	\caption{Loss balance $\lambda$}
\end{subfigure}%
\begin{subfigure}{0.27\textwidth}
\centering
    \includegraphics[width=1\linewidth]{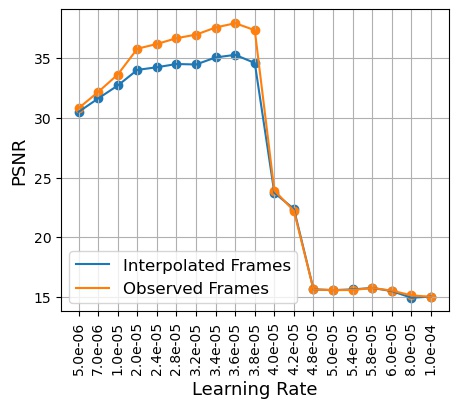}
	\caption{Learning Rate}
\end{subfigure}%
\begin{subfigure}{0.31\textwidth}
	\centering
    \includegraphics[width=1\linewidth]{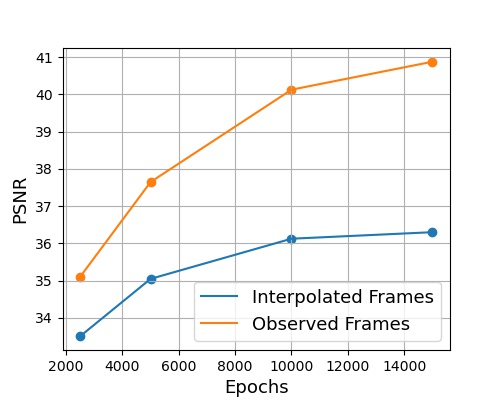}
    \caption{Epochs}
\end{subfigure}

\begin{subfigure}{0.3\textwidth}
\centering
    \includegraphics[width=1\linewidth]{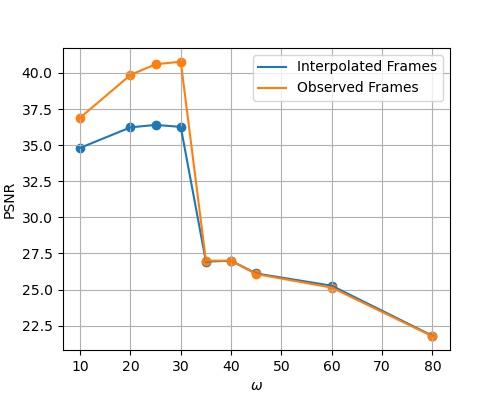}
    \caption{$\omega$}
    \label{fig_abl_omega}
\end{subfigure}%
\begin{subfigure}{0.3\textwidth}
\centering

\includegraphics[width=1\linewidth]{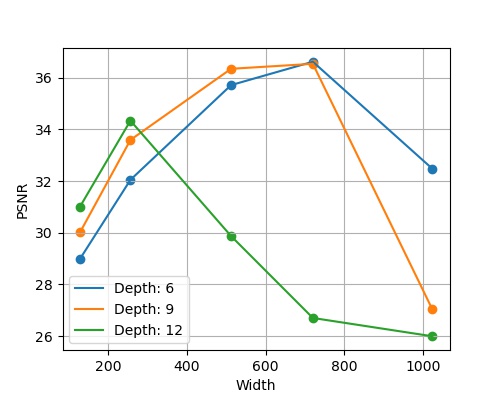}
    \caption{Width and Depth}
\end{subfigure}
\caption{Impact of our method's main parameters. Plots from (a) to (d)
show both the observed and interpolated frames PSNR
while plot (e) only shows the interpolated frames PSNR.}
\label{fig_abl}
\end{figure}

Figure \ref{fig_abl} summarizes the impact of the main parameters of our method.
In (a) we observe a trade-off between the observed and interpolated frames quality in the low $\lambda$ ranges.
The quality of interpolated frames peaks at $\lambda=0.12$, beyond which point the interpolated frames quality is limited
by the quality of the fit to the observed frames, in a similar way to the classical overfitting/underfitting trade-off.
However, it should be noted that this trade-off differs widely depending on the SIREN's width.
Indeed, as we will show in Section \ref{sec_video_fit},
the OF constraints actually improves the fit to observed frames for narrow models.
In (b) and (c) we observe that both higher learning rates and longer fitting times
improve both observed and interpolated frames.
The learning rate is limited in amplitude by instabilities of the optimization procedure,
while the fitting time is limited by practical time constraints.
Large $\omega$ (d) also improve the accuracy up to 30,
after which instabilities in the optimization see the accuracy drop abruptly.
Width and depth (e) show interesting co-dependencies:
Increasing width improves interpolation up to a peak after which it degrades.
The peak width gets smaller with increasing depth.

Based on these experiments, our final results, as reported in Table 1 were computed with a SIREN model with depth 6, width 720 and $\omega=25$.
We used $\lambda = 0.12$ for the loss, and optimized using Adam with a maximum learning rate of 3.6e-5 during 15k epochs.

\subsection{Qualitative analysis}
\label{sec_qua}

\begin{figure}[h]
\centering
\begin{subfigure}{0.14\textwidth}
	\centering
    \includegraphics[width=1\linewidth]{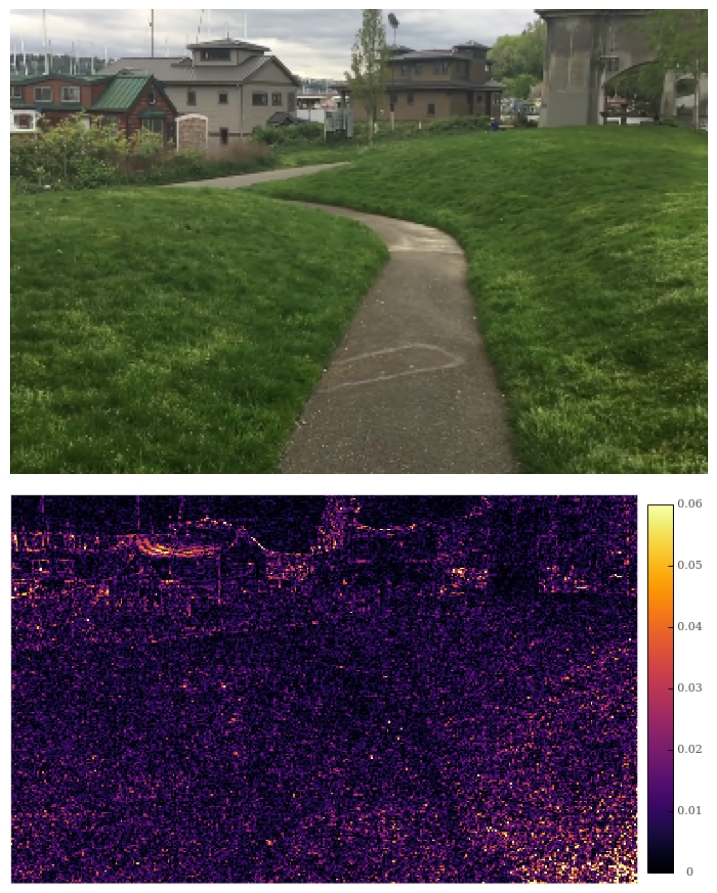}
\end{subfigure}%
\begin{subfigure}{0.14\textwidth}
	\centering
    \includegraphics[width=1\linewidth]{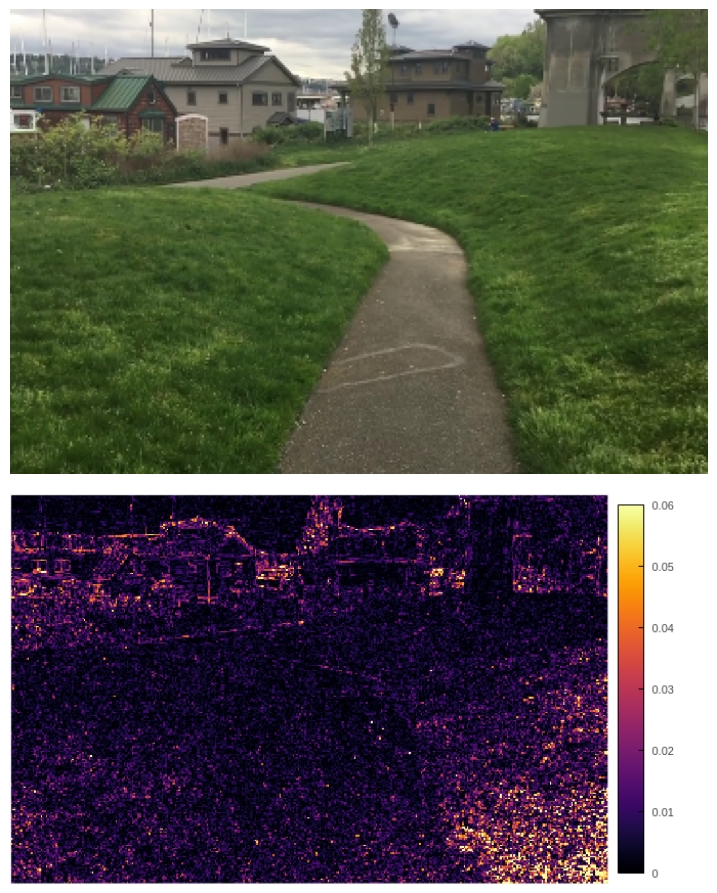}
\end{subfigure}%
\begin{subfigure}{0.14\textwidth}
	\centering
    \includegraphics[width=1\linewidth]{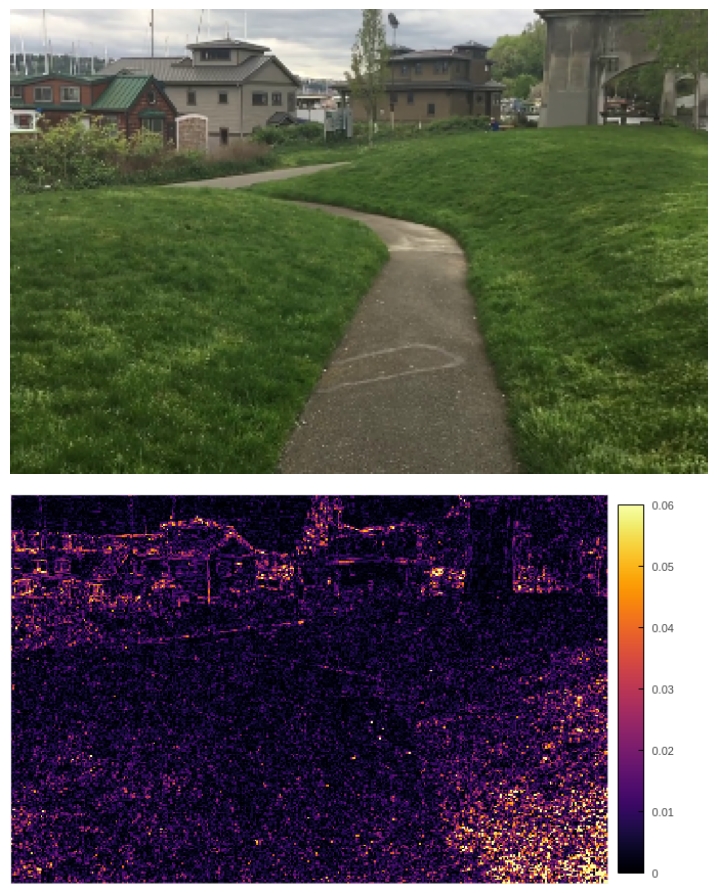}
\end{subfigure}%
\begin{subfigure}{0.14\textwidth}
	\centering
    \includegraphics[width=1\linewidth]{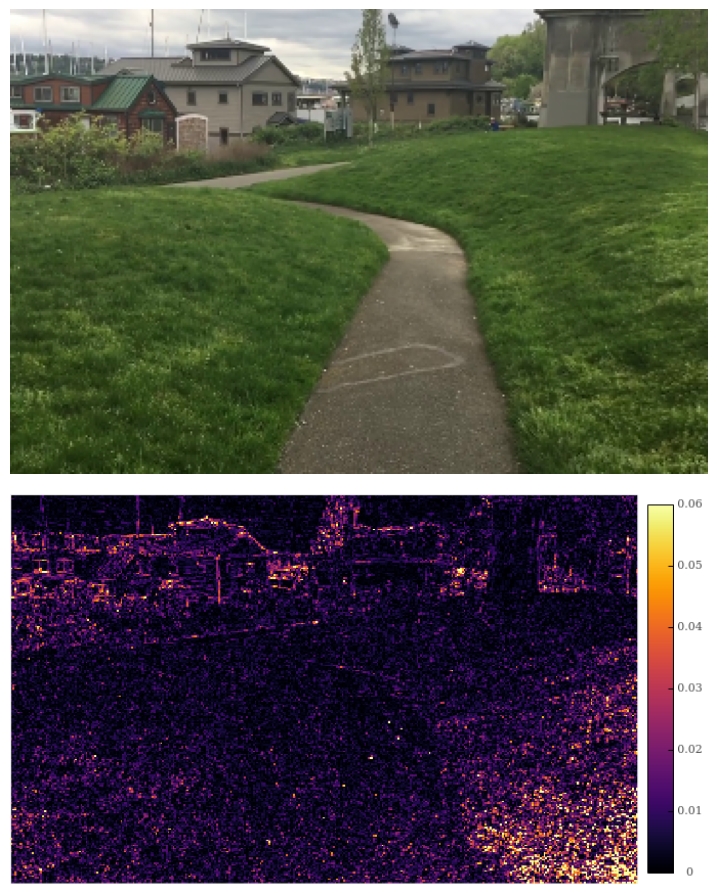}
\end{subfigure}%
\begin{subfigure}{0.14\textwidth}
	\centering
    \includegraphics[width=1\linewidth]{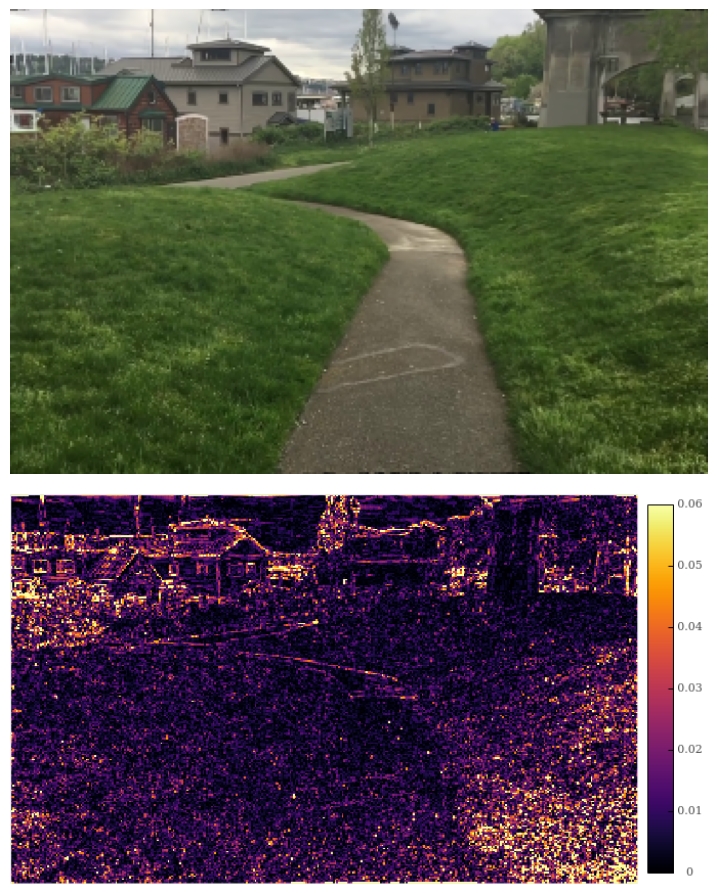}
\end{subfigure}%
\begin{subfigure}{0.14\textwidth}
	\centering
    \includegraphics[width=1\linewidth]{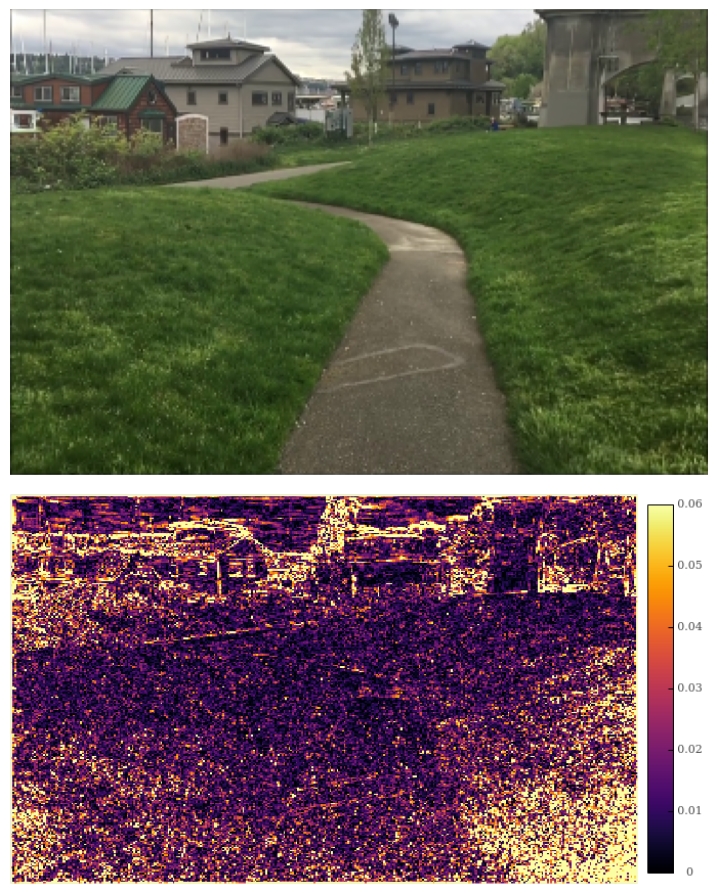}
\end{subfigure}%
\begin{subfigure}{0.14\textwidth}
	\centering
    \includegraphics[width=1\linewidth]{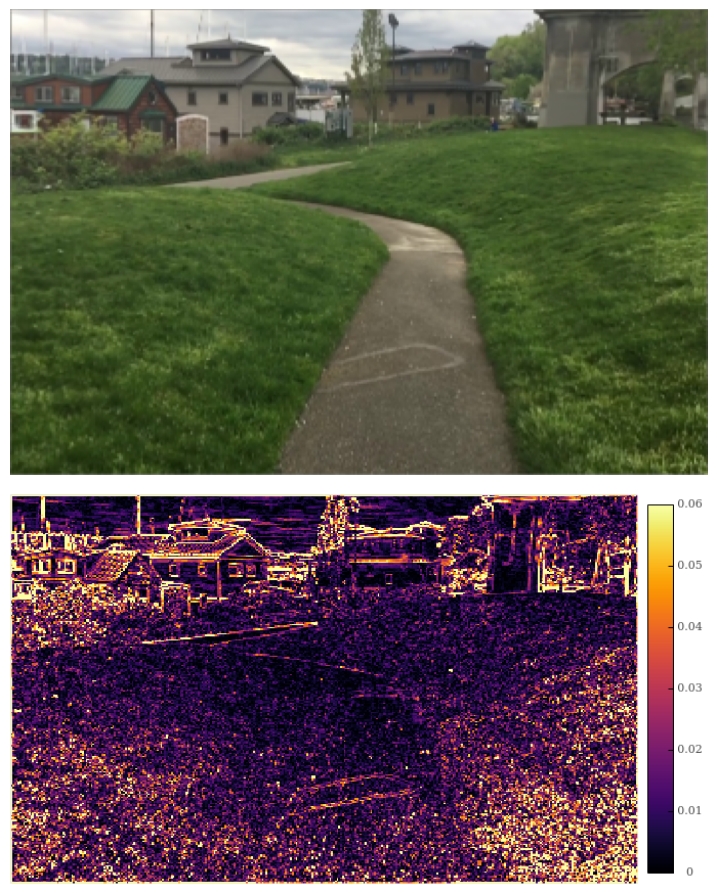}
\end{subfigure}
\begin{subfigure}{0.14\textwidth}
	\centering
    \includegraphics[width=1\linewidth]{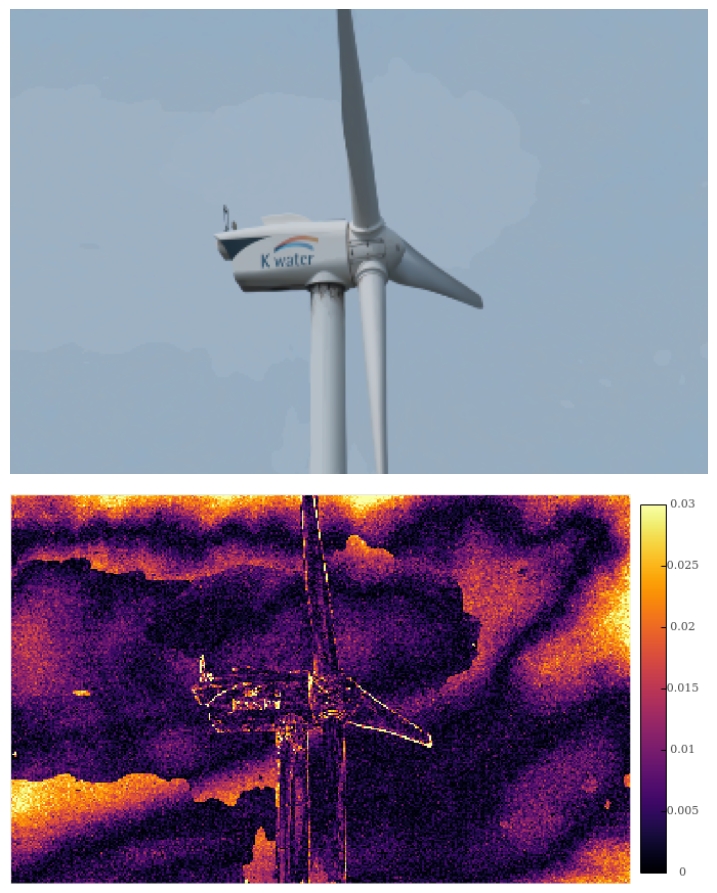}
    \caption{Ours}
\end{subfigure}%
\begin{subfigure}{0.14\textwidth}
	\centering
    \includegraphics[width=1\linewidth]{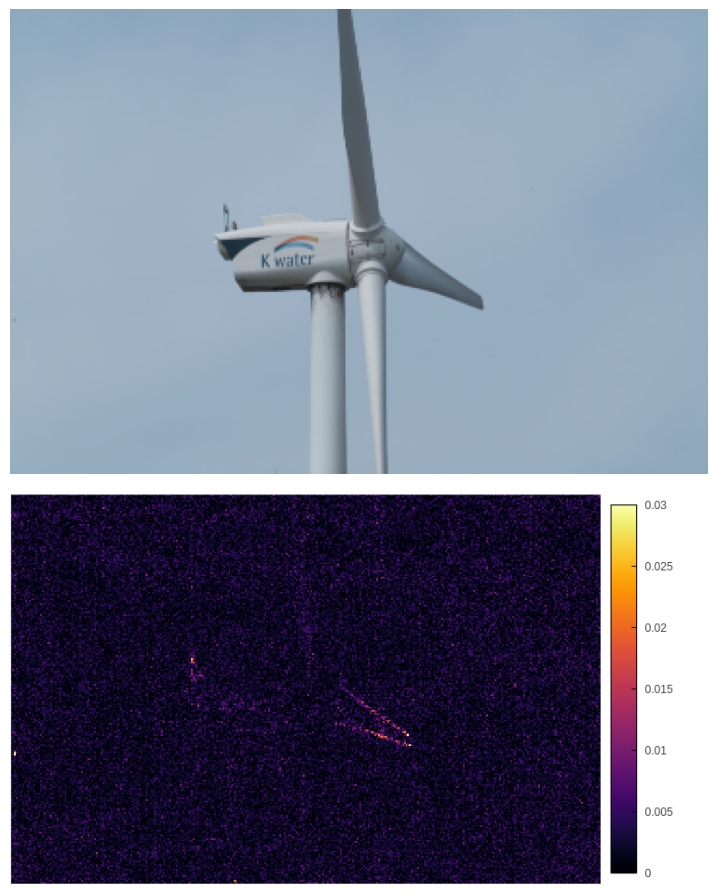}
    \caption{ABME}
\end{subfigure}%
\begin{subfigure}{0.14\textwidth}
	\centering
    \includegraphics[width=1\linewidth]{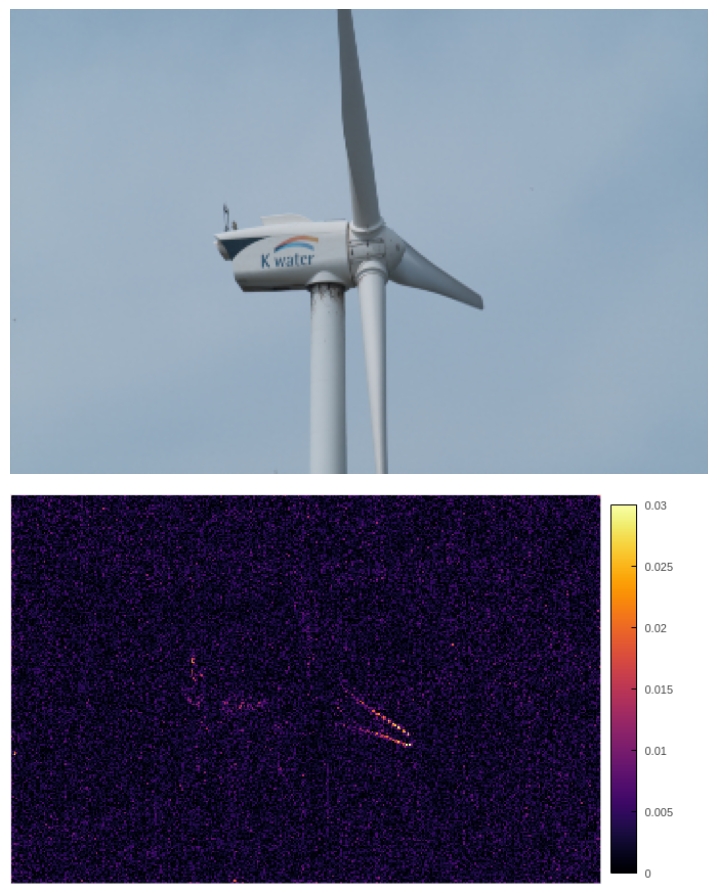}
    \caption{FILM}
\end{subfigure}%
\begin{subfigure}{0.14\textwidth}
	\centering
    \includegraphics[width=1\linewidth]{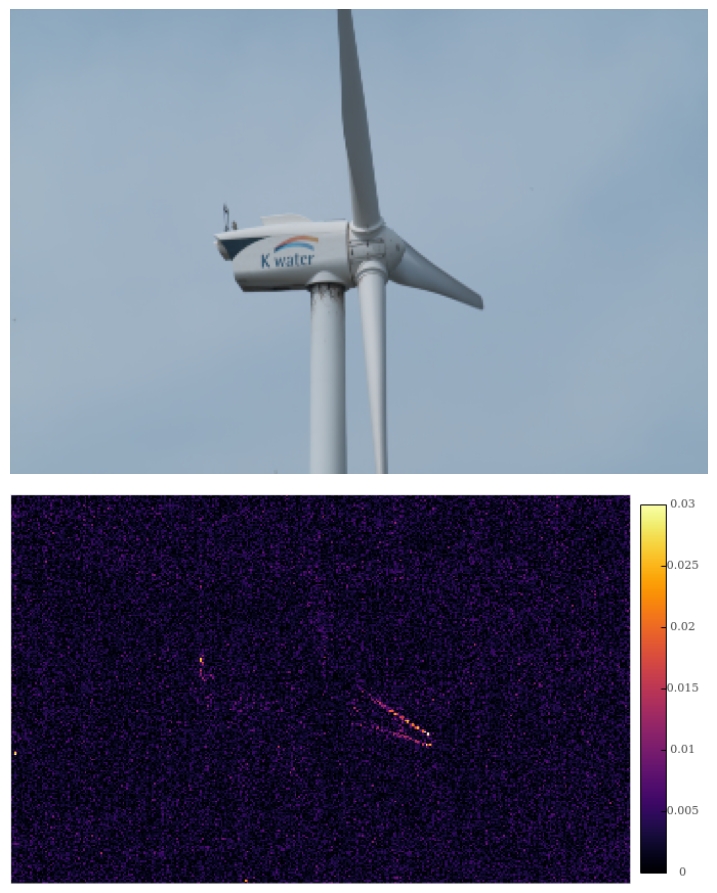}
    \caption{AdaCof}
\end{subfigure}%
\begin{subfigure}{0.14\textwidth}
	\centering
    \includegraphics[width=1\linewidth]{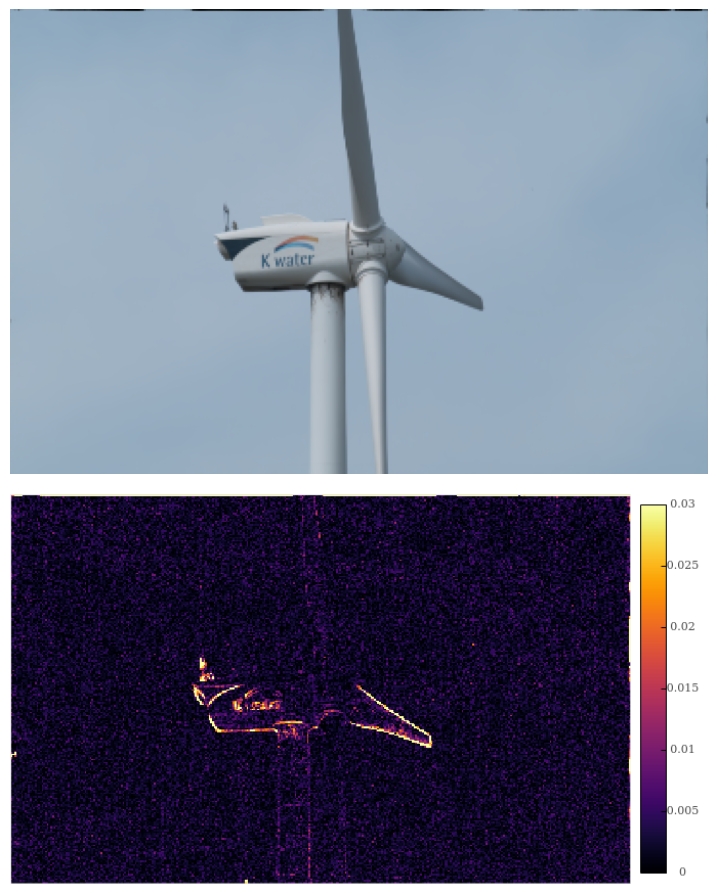}
    \caption{BMBC}
\end{subfigure}%
\begin{subfigure}{0.14\textwidth}
	\centering
    \includegraphics[width=1\linewidth]{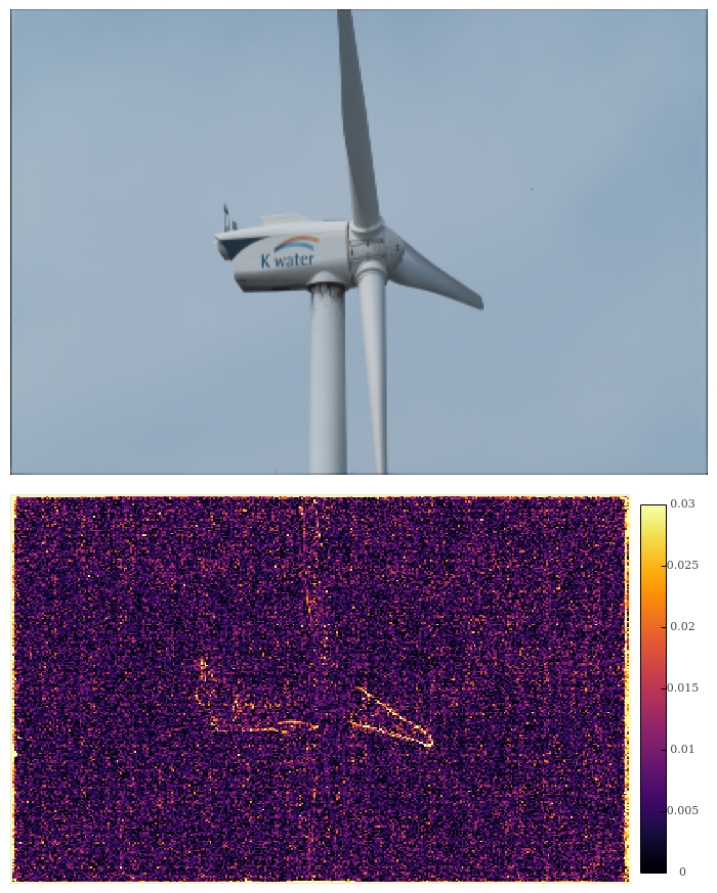}
    \caption{RRIN}
\end{subfigure}%
\begin{subfigure}{0.14\textwidth}
	\centering
    \includegraphics[width=1\linewidth]{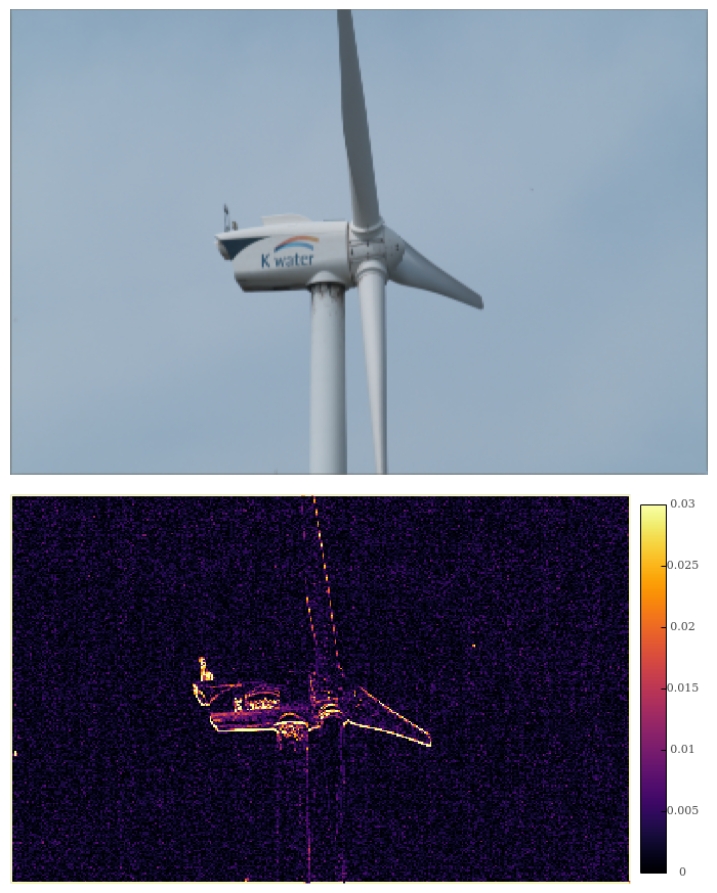}
    \caption{SloMo}
\end{subfigure}
\caption{
Small Motion Video Qualitative Analysis.
The interpolated frame results are shown above their residual heat map.
The upper frames show a successfully interpolated frames,
the lower one shows a rare failure case.}
\label{fig_small}
\end{figure}

\begin{figure}[h]
\centering
\begin{subfigure}{0.2\textwidth}
	\centering
    \includegraphics[width=1\linewidth]{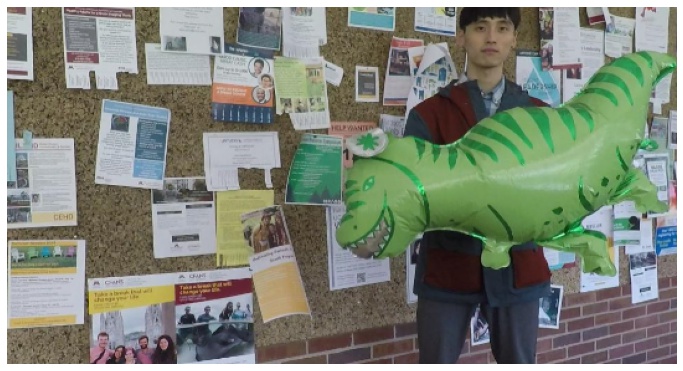}
\end{subfigure}%
\begin{subfigure}{0.2\textwidth}
	\centering
    \includegraphics[width=1\linewidth]{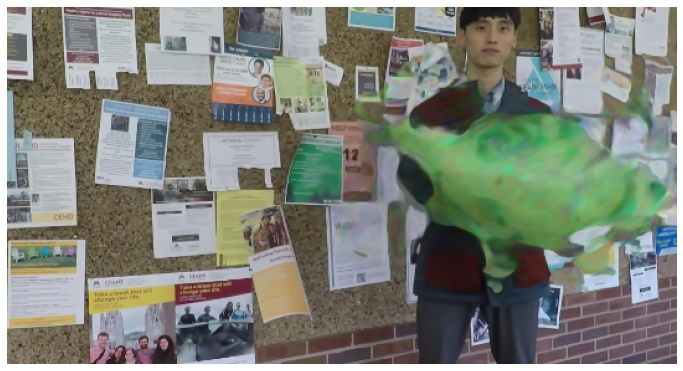}
\end{subfigure}%
\begin{subfigure}{0.2\textwidth}
	\centering
    \includegraphics[width=1\linewidth]{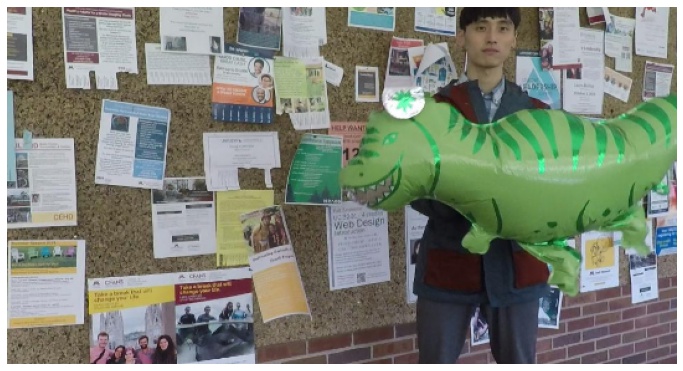}
\end{subfigure}%
\begin{subfigure}{0.2\textwidth}
	\centering
    \includegraphics[width=1\linewidth]{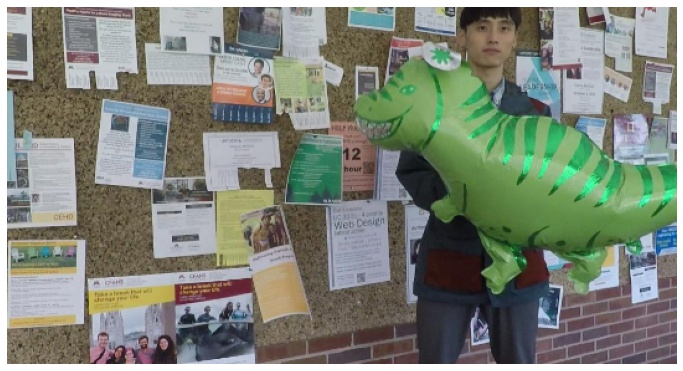}
\end{subfigure}%
\begin{subfigure}{0.2\textwidth}
	\centering
    \includegraphics[width=1\linewidth]{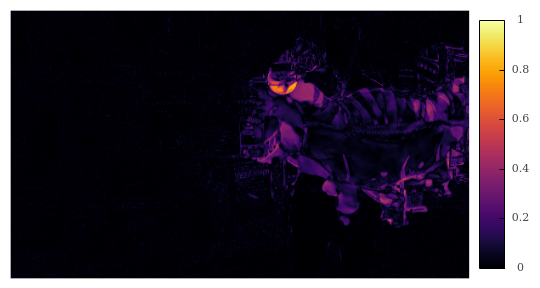}
\end{subfigure}
\begin{subfigure}{0.2\textwidth}
	\centering
    \includegraphics[width=1\linewidth]{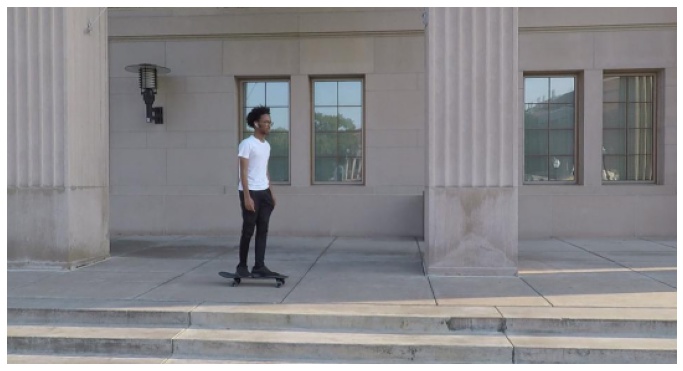}
    \caption{$v(x,y,0)$}
\end{subfigure}%
\begin{subfigure}{0.2\textwidth}
	\centering
    \includegraphics[width=1\linewidth]{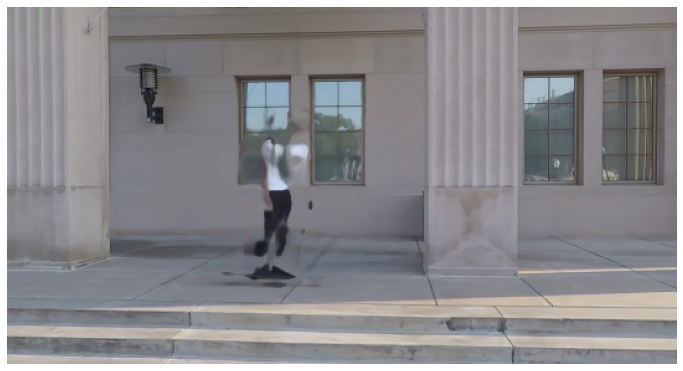}
    \caption{$f_{\theta}(x,y,0.5)$}
\end{subfigure}%
\begin{subfigure}{0.2\textwidth}
	\centering
    \includegraphics[width=1\linewidth]{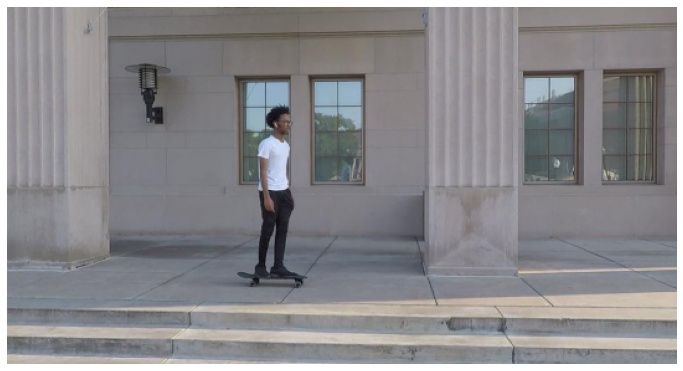}
    \caption{$v(x,y,0.5)$}
\end{subfigure}%
\begin{subfigure}{0.2\textwidth}
	\centering
    \includegraphics[width=1\linewidth]{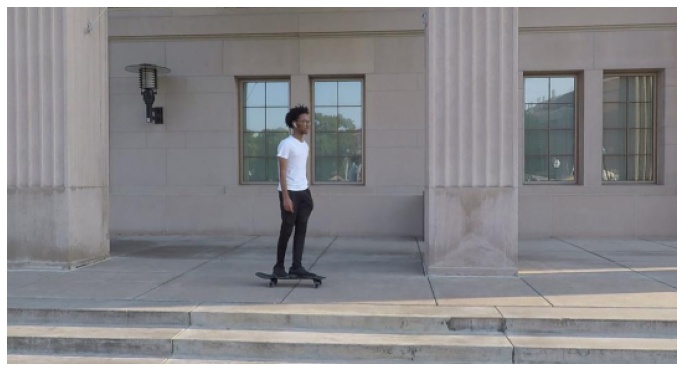}
    \caption{$v(x,y,1)$}
\end{subfigure}%
\begin{subfigure}{0.2\textwidth}
	\centering
    \includegraphics[width=1\linewidth]{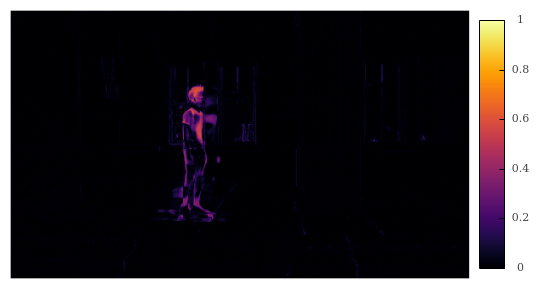}
    \caption{Residual Error}
\end{subfigure}
\caption{Large Motion Video Qualitative Analysis}
\label{fig_large}
\end{figure}

Figures \ref{fig_small} and \ref{fig_large} provide a qualitative illustration to the results presented in Section \ref{sec_sota}.
The upper frame in Figure \ref{fig_small} shows that our method tends to outperform other methods on videos with limited motion range.
In particular it seems to better catch high spatial frequency regions (grass, sharp edges of the building).
In contrast, large motion as illustrated in Figure \ref{fig_large} shows ghosting effects that the OF regularization is not able to address.

The lower part of Figure \ref{fig_small} shows a rare failure case of our method on limited motion ranges:
some artificial stain-like patterns appear in the sky background,
suggesting additional care may be needed especially in low frequency regions.
Despite this rare exception, the overall quality of interpolation
on limited motion range videos performs on par with the best existing methods.

\subsection{Video fitting}

\begin{figure}[h]
\centering
\includegraphics[width=1\textwidth]{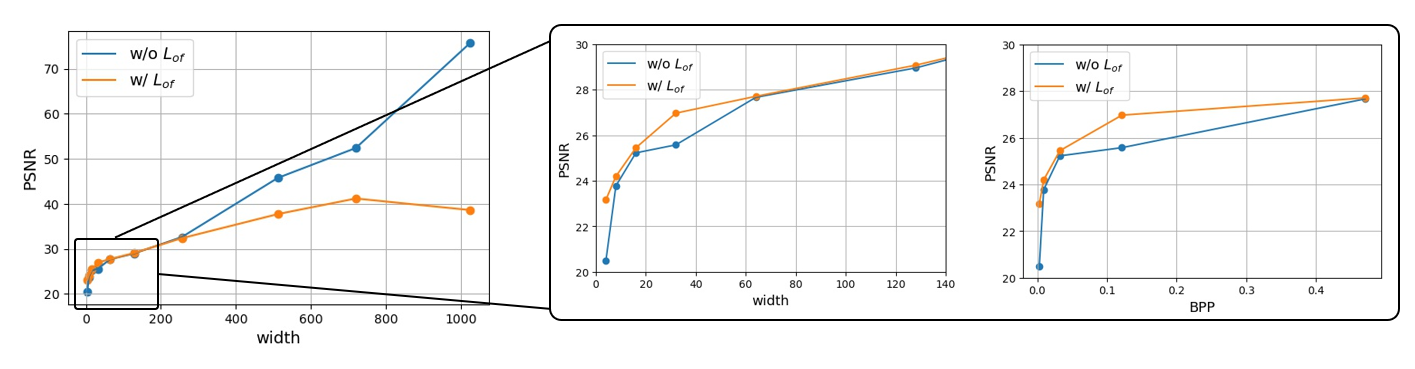}
\caption{Evolution of the \textbf{observed} frames PSNR with depth, with and without OF regularization.
Left: Trend from very narrow to very wide models.
Right: Zoom on the low width regime with the x axis expressed either in number of neurons or corresponding Bits Per Pixel measure.}
\label{fig_video_fit}
\end{figure}

Figure \ref{fig_video_fit} shows an unexpected side-effect of the OF regularization observed for narrow networks.
As $\mathcal{L}_{obs}$ explicitly maximizes the PSNR of observed frames,
we expected the addition of the $\mathcal{L}_{of}$ term to negatively impact the PSNR of observed frames,
especially for capacity-limited SIREN which should have to compromise between satisfying both loss terms.
It turns out that, for width up to 50, optimizing the SIREN with the additional OF constraint
actually improves the fit to observed frames.

\label{sec_video_fit}

Although a complete investigation of this phenomenon is out of the scope of this work,
we highlight how this observation may prove interesting for future works:
From a practical standpoint, improving the fit of low-capacity INR
is the key challenge towards practical INR video compression.
It remains to be seen whether this phenomenon can be replicated on more practical architectures (i.e. \cite{chen2021nerv}).
From a theoretical standpoint, increasing width has been shown to help
optimization by alleviating second order effects \cite{liu2020linearity}
and guarantee convergence of gradient descent to global minima \cite{du2018gradient}.
As the understanding of gradient descent dynamics in the high curvature low width setting
is currently an elusive question, understanding how the OF constraint
helps optimization may provide useful insights into gradient descent dynamics in narrow models.

\section{Current Limitations and Future Work}
\label{sec_lim}

While our method does reach state of the art interpolation results on limited motion ranges,
this work is not meant to deliver a production ready VFI system,
which would require the ability to interpolate high resolution and large motion range videos.
Instead, we aim to provide actionable insights for future works on both VFI and INR to integrate and build upon.
Towards that goal, we discuss below what we see as the three main limitations of our method in its current form,
and possible ways to address these limitations.

\textbf{Slow optimization process.}
Fitting 20 frames of a video at $240 \times 360$ resolution currently takes 15 hours on a 4$\times$2080Ti GPU using Pytorch.
This computation time is an important drawback as it limits our ability to process full resolution video,
as well as to explore different hyper parameters and variations of the method within realistic times.
We expect advances in INR optimization to be very beneficial to this line of research.
Given recent successes of INR in signal compression \cite{zhang2021implicit}\cite{dupont2021coin} \cite{dupont2022coin++} \cite{park2019deepsdf} \cite{mescheder2019occupancy} \cite{chen2021nerv},
we hopefully expect to see such development in the near future.

\textbf{Reliance on trained optical flow model.}
SIREN models allow us to apply the optical flow on the exact derivatives of the signal,
bypassing the heuristics of classical methods without relying on machine learning.
The optical flow we use, however, is given by a ML model trained on discrete representations, which raises two problems:
it is subject to generalization errors,
and is subject to finite difference errors such as occlusions.
Bypassing this reliance on ML-based OF using alternative constraints on the exact derivatives
of the representation is another interesting way forward.

\textbf{Inability to interpolate large motion range videos.}
In its current form, we only apply the optical flow constraint on the observed frames of the video.
This has proven sufficient to reach state-of-the art on limited motion ranges, but is not sufficient for large motions.
A promising axis of improvement would be to apply additional constraints to the interpolated frames
(e.g. for intra-frame time indices $t=0.5$).
Possible regularization methods may include constraints on intra-frame texture, as proposed in recent works \cite{reda2022film},
or interpolated optical flows, which may prevent the ghosting effects illustrated in Figure \ref{fig_large}.

\section{Conclusion}
\label{sec_conc}

In this paper, we have shown that SIREN representations of videos
can be constrained to satisfy the OF constraint equation in their exact derivatives.
We have seen that OF-constrained SIREN reach state of the art VFI on limited motion ranges,
without relying on ML based residual flow and interpolation.
We have also shown that the OF constraint not only allows SIREN to generate intermediate frames,
but can also improve the ability of narrow SIREN to fit observed frames.
We have discussed the limitations of our approach in its current form and outlined
potentially impactful way forwards for future research.

\bibliographystyle{abbrv}
\bibliography{references}

\end{document}